\documentclass{article}
\usepackage{iclr2027_conference,times}
\usepackage[utf8]{inputenc}
\usepackage[T1]{fontenc}
\usepackage{graphicx,booktabs,amsfonts,amsmath}
\usepackage{microtype,xcolor,xspace,wrapfig,multirow,enumitem}
\usepackage[hidelinks]{hyperref}
\usepackage{etoolbox}
\usepackage{placeins}
\makeatletter
\ifdefined\pdfrunninglinkoff
  \AddToHook{shipout/before}{%
    \setbox\ShipoutBox\vbox{\pdfrunninglinkoff\box\ShipoutBox}}
  \patchcmd{\@outputpage}{\box\@outputbox}
    {\pdfrunninglinkon\box\@outputbox\pdfrunninglinkoff}
    {}{\PackageError{manuscript}{Could not isolate body links}{Check the output routine.}}
\fi
\makeatother
\usepackage{url}
\usepackage{tikz}
\usetikzlibrary{arrows.meta,positioning}

\newcommand{\ASL}{\ensuremath{\text{ASL}}\xspace}
\newcommand{\CSL}{\ensuremath{\text{CSL}}\xspace}
\newcommand{\DGS}{\ensuremath{\text{DGS}}\xspace}
\newcommand{\ASLfull}{\ensuremath{\text{American Sign Language (\ASL)}}\xspace}
\newcommand{\CSLfull}{\ensuremath{\text{Chinese Sign Language (\CSL)}}\xspace}
\newcommand{\DGSfull}{\ensuremath{\text{German Sign Language (\DGS)}}\xspace}

\newcommand{\SStoSS}{\ensuremath{\text{sign}\!\leftrightarrow\!\text{sign}}\xspace}
\newcommand{\TS}{\ensuremath{\text{T2S}}\xspace}
\newcommand{\ST}{\ensuremath{\text{S2T}}\xspace}
\newcommand{\StwoS}{\ensuremath{\text{S2S}}\xspace}
\newcommand{\bidir}{ \ensuremath{\leftrightarrow} }

\newcommand{\soke}{\ensuremath{\text{\textsc{SOKE}}}\xspace}

\newcommand{\qwen}{Qwen\xspace}
\newcommand{\qwenthree}{Qwen3\xspace}
\newcommand{\qwenbackbone}{Qwen3-0.6B\xspace}
\newcommand{\qwenjudge}{Qwen3-8B\xspace}
\newcommand{\unisign}{Uni-Sign\xspace}
\newcommand{\geosign}{Geo-Sign\xspace}
\newcommand{\mmslt}{MMSLT\xspace}
\newcommand{\translategemma}{TranslateGemma\xspace}
\newcommand{\slt}{SLT\xspace}
\newcommand{\sbert}{S-BERT\xspace}
\newcommand{\smplx}{\ensuremath{\text{SMPL-X}}\xspace}

\newcommand{\howtosign}{\ensuremath{\text{How2Sign}}\xspace}
\newcommand{\csldaily}{\ensuremath{\text{CSL-Daily}}\xspace}
\newcommand{\phoenix}{\ensuremath{\text{Phoenix-2014T}}\xspace}

\newcommand{\inantest}{\ensuremath{\text{Inan-Test}}\xspace}
\newcommand{\inancleaned}{\ensuremath{\text{Inan-Strict}}\xspace}

\newcommand{\mpjpe}{MPJPE\xspace}
\newcommand{\pampjpe}{PA-MPJPE\xspace}
\newcommand{\dtwpa}{\ensuremath{\text{\textsc{DTW-PA-MPJPE}}}\xspace}

\newcommand{\bleufour}{\ensuremath{\text{BLEU-4}}\xspace}

\newcommand{\BT}{\ensuremath{\text{BT}}\xspace}

\newcommand{\signseq}{\boldsymbol{y}}
\newcommand{\textseq}{\boldsymbol{x}}
\newcommand{\prefixseq}{\boldsymbol{c}}
\newcommand{\modelparams}{\boldsymbol{\theta}}

\title{Direct Translation between Sign Languages}
\author{
Zetian Wu \quad
Bowen Xie \quad
Wuyang Meng \quad
Milan Gautam \quad
Stefan Lee \quad
Liang Huang \quad\\
Oregon State University \\
{\footnotesize\texttt{\{wuzet, xiebo, mengwu, gautammi, leestef, liang.huang\}@oregonstate.edu}}
}
\iclrfinalcopy
\makeatletter
\patchcmd{\@maketitle}{Published as a conference paper at ICLR 2027}{Preprint}
  {}{\PackageError{preprint}{Could not replace conference header}{Check the template.}}
\makeatother
\begin{document}
\maketitle

\begin{abstract}
Sign language translation has made substantial progress between sign
and spoken languages, while translation across sign languages remains
less explored. Translating directly between sign languages could
support communication across signing communities without requiring
a shared written language. A cascade of sign-to-text, spoken-language
translation, and text-to-sign models offers one route, but can
propagate intermediate errors and requires three sequential translation
stages. We develop direct sign-to-sign translation, whose training is
limited by the scarcity of parallel signing across languages.
To address this obstacle, we adapt back-translation to construct
cross-lingual pairs from existing text--sign corpora: the source
signing is synthesized through a text bridge, while the target
remains the gold sign from the original corpus. Using these pairs, we jointly train
a single \qwenthree{}-based model for text-to-sign and sign-to-sign translation.
The latter generates target signing directly from source signing
without an intermediate transcript. Experiments cover six directions
among American, Chinese, and German Sign Language, using back-translation
pairs and existing cross-lingual sign pairs.
On synthetic sources, direct translation improves \bleufour{} in five
directions and lowers overall motion error relative to the cascade.
On existing test pairs, it improves \bleufour{} over
the cascade by 1.91--3.87 points across all six directions.
\end{abstract}

\section{Introduction}
\label{sec:intro}

\begin{figure}[htbp]
  \centering
  \includegraphics[width=\linewidth]{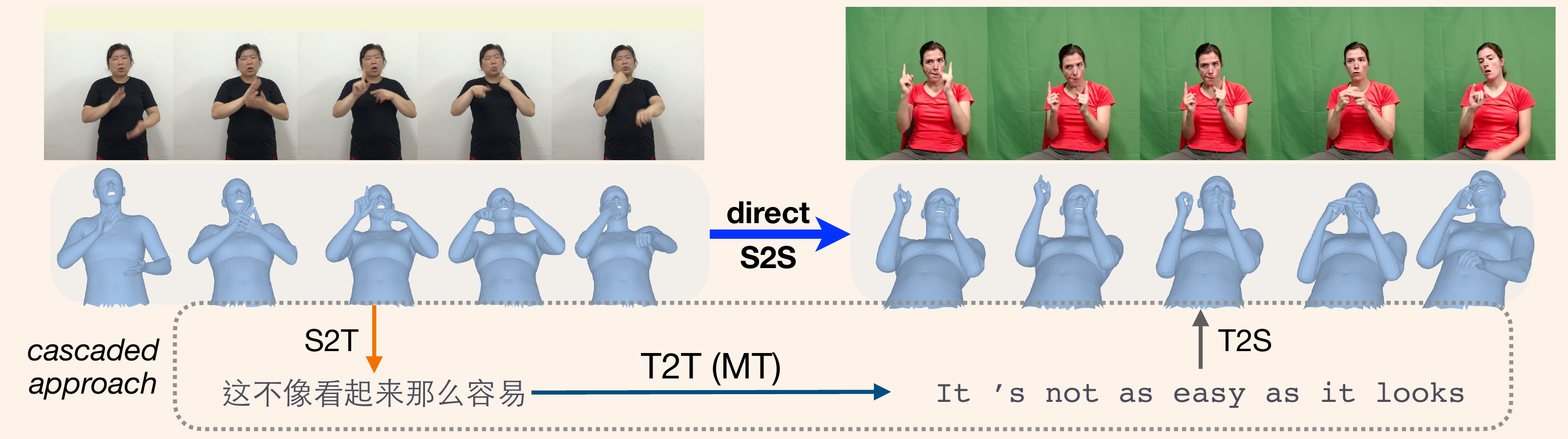}
  \caption{\textbf{Direct sign-to-sign translation.} Illustration of translation from \CSL to \ASL. The direct route (blue arrow) maps source signing to target signing without intermediate text. The cascaded route passes through source-language text, spoken-language MT, and target-language text before generating target signing.}
  \label{fig:teaser}
\end{figure}

Deaf and hard-of-hearing signers from different countries face barriers to conversing in their native sign languages. Like spoken languages, sign languages have distinct grammars and vocabularies: \ASLfull, \CSLfull, and \DGSfull are different languages, and hundreds more are in active use worldwide~\citep{yin2021including}. Such a conversation may rely on human interpreters, whose availability is limited, or on written text in a spoken language that the signers may not share. The latter route takes the conversation out of sign and can lose the spatial grammar, classifier predicates and prosody that signed dialogue relies on~\citep{yin2021including, decoster2023machine}. A system that translates directly between sign languages---taking one signer's clip in, returning an equivalent clip in another, with no detour through spoken-language text---could help address this barrier (Figure~\ref{fig:teaser}).

The most natural way is a \emph{cascade}: chain a sign-to-text (\ST) model~\citep{camgoz2018slt, camgoz2020slt, lin2023glossfree, yin2020stmctransformer} with a spoken-language MT system~\citep{liu2020mbart, nllb2022} and a text-to-sign (\TS) generator~\citep{stoll2018slp, saunders2020progressive, saunders2021mixed, zelinka2020neural, soke2025}. This route reuses well-studied components, but it compounds errors across three stages, requires three sequential translation stages per query, and discards visual information not captured in the intermediate text.

A direct sign-to-sign (\StwoS) model bypasses this cascade, removing both the intermediate text bottleneck and the need to run three translation models in sequence (as in direct \emph{speech-to-speech translation}~\citep{translatotron2019, translatotron2-2022, audiopalm2023} in the spoken modality), but it is held back by the lack of parallel \StwoS data.
The pioneering direct \StwoS work of \citet{inan2025align} automatically aligned three pairwise corpora (\ASL\bidir\CSL, \ASL\bidir\DGS, \CSL\bidir\DGS) and trained a pose-to-pose Transformer based on~\citet{saunders2020progressive}.
Their constructed data have two major limitations: \emph{scale} and \emph{alignment quality}.
The training sets contain only 317--2,274 pairs per language pair. The authors themselves describe the construction as a ``valuable yet noisy training set,'' acknowledging that some paired clips do not share the same meaning. In their limitations, they further explain that domain gaps between corpora make it impractical to identify equivalent clips from transcripts for longer, more complex meanings.
Their reported \bleufour, computed on text recovered from generated signing, is zero in two directions and at most 6.81 elsewhere.\footnote{These automatic scores also depend on the sign-to-text evaluator and tokenization, as discussed by \citet{inan2025align}.}
Thus, direct \StwoS still needs a way to obtain substantially more paired supervision without relying on semantic overlap between independently collected corpora.

We address this data bottleneck by importing
\emph{back-translation} (\BT)~\citep{sennrich2016bt, edunov2018bt}\footnote{Throughout this paper, \emph{back-translation} refers to constructing synthetic-source training pairs, following \citet{sennrich2016bt}.},
a widely used technique for low-resource neural machine translation (NMT), into the sign modality: for each gold (text, sign) pair in a monolingual corpus, we translate the text into another spoken language, feed the result to a fixed \TS model to produce a synthetic source-sign clip, and pair that clip with the original gold sign as the target---yielding a large-scale parallel \StwoS training corpus (\S\ref{sec:bt}). For example, given an English--\ASL pair $(\textseq^{\mathrm{en}}, \signseq^{\mathrm{ASL}})$ from \howtosign, we translate $\textseq^{\mathrm{en}}$ into Chinese $\hat{\textseq}^{\mathrm{zh}}$ via MT,
and then use the \TS model to generate a synthetic \CSL clip $\hat{\signseq}^{\mathrm{CSL}}$ from $\hat{\textseq}^{\mathrm{zh}}$; finally, we pair $(\hat{\signseq}^{\mathrm{CSL}}, \signseq^{\mathrm{ASL}})$ as a \CSL$\to$\ASL training instance. This construction does not require a matching source clip to exist in another corpus, and every target is a gold clip from the original corpus. Sign-side \BT has previously been used in sign-to-text translation to augment sign--text training pairs~\citep{zhou2021back, moryossef2021augmentation}; here we use it to construct parallel training data across sign languages. Using our separately trained sign tokenizer, we train a single \qwenthree{}-based model~\citep{qwen3-2025} jointly on \TS and \StwoS, sharing target-sign generation across natural text--sign pairs and our synthetic-source, gold-target pairs (\S\ref{sec:model}).

Concretely, this paper makes three main contributions:
\begin{itemize}
    \item We construct a large-scale parallel \SStoSS training corpus of 112,324 ordered pairs by porting back-translation~\citep{sennrich2016bt, edunov2018bt} into the sign modality: a \TS model produces synthetic source clips from translated texts, which we pair with gold target clips from monolingual sign corpora (\S\ref{sec:bt}).
    \item While there exists a prior set of cross-lingual \SStoSS dataset \citep{inan2025align}, it was aligned automatically and noisy by their own report; we extract a stricter diagnostic subset using an LLM judge, Sentence-BERT~\citep{reimers2019sbert}, and manual transcript screening to reduce semantic mismatches in the paired references (\S\ref{sec:dataset}).
    \item Direct translation achieves higher \bleufour{} than the cascade in all six \inantest{} directions (\S\ref{sec:experiment}). On synthetic sources, it outperforms the cascade in five directions on both BLEU and motion error, except DGS$\to$ASL. It is $2.6\times$ faster than the cascade's summed stage times.
\end{itemize}

\section{Synthetic Sign-to-Sign Corpus via Back-Translation}
\label{sec:bt}

\begin{figure}[htbp]
  \centering
  \includegraphics[width=\linewidth]{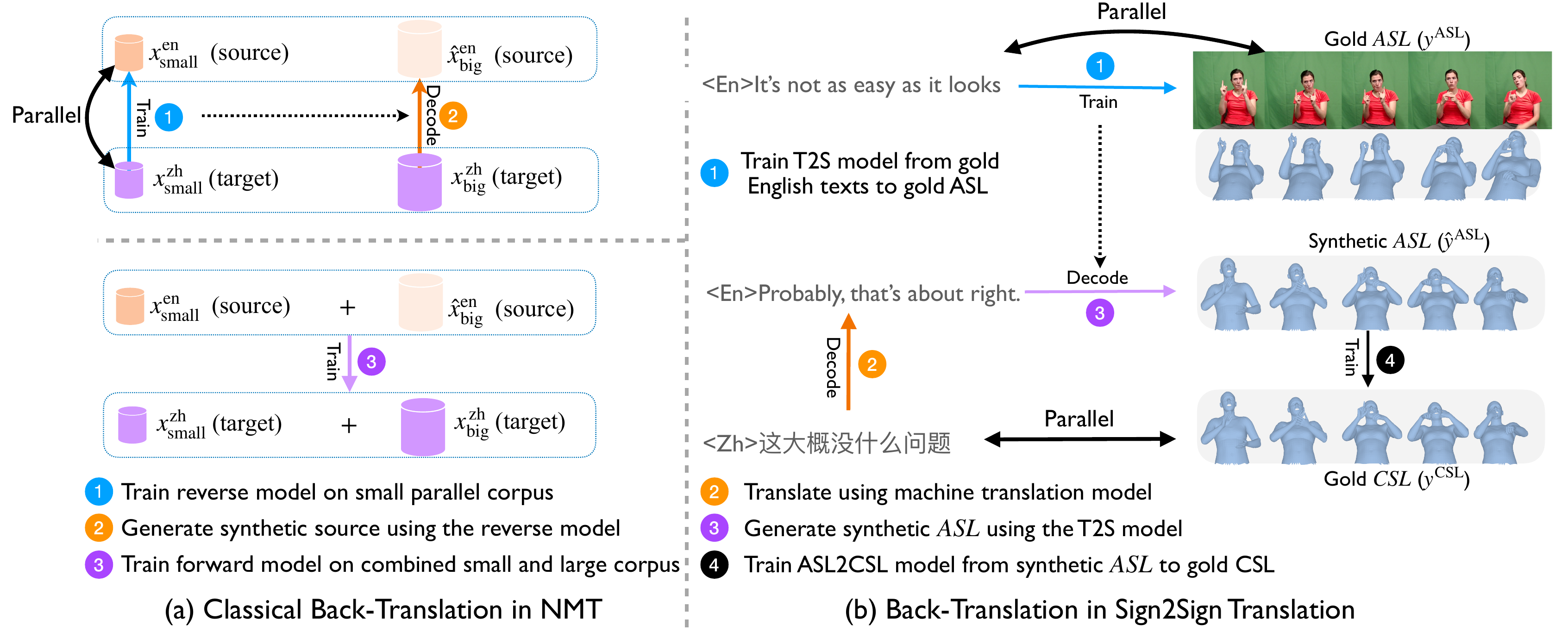}
  \caption{\textbf{Source-sign synthesis via a text bridge.}
  \textbf{(a)} Standard NMT synthesizes sources from gold
  target sentences~\citep{sennrich2016bt}.
  \textbf{(b)} We translate a gold CSL clip's transcript into English,
  generate an ASL source with \TS{}, and pair it with the original CSL target.}
  \label{fig:bt}
\end{figure}

Large parallel \SStoSS corpora are scarce, so we use back-translation
to construct synthetic-source, gold-target \StwoS pairs from text--sign
corpora (Figure~\ref{fig:bt}).

\subsection{Classical Back-Translation in NMT}
\label{sec:bt:nmt}
In NMT, back-translation expands scarce parallel data using abundant
monolingual target-language text, without new human
translations~\citep{sennrich2016bt,edunov2018bt}.

Let $\textseq^{l}$ and $\textseq^{l'}$ denote target- and source-language sentences.
A reverse model trained on the small parallel corpus maps each gold
monolingual target $\textseq^{l}$ to a synthetic source $\hat{\textseq}^{l'}$.
The pairs $(\hat{\textseq}^{l'},\textseq^{l})$ augment the original parallel data for
the forward model (Figure~\ref{fig:bt}(a)). For example, to train a
German-to-English model, we generate synthetic German sentences from
monolingual English text using an English-to-German model. Each generated sentence
is paired with its original English target.
We likewise synthesize source signing while retaining gold target clips
(Figure~\ref{fig:bt}(b)).

\subsection{Adapting Back-Translation to Sign-to-Sign Translation}
\label{sec:bt:s2s}
For sign language $s$, let $\signseq^{s}$ denote a sign sequence paired
with a transcript $\textseq^{l(s)}$ in its associated spoken language $l(s)$. We use \howtosign for English--\ASL~\citep{duarte2021how2sign},
\csldaily for Chinese--\CSL~\citep{zhou2021back}, and \phoenix for
German--\DGS~\citep{camgoz2018slt}. These corpora provide paired transcripts
and clips for each sign language, but do not directly provide parallel sign
pairs across languages.

For each gold text--sign pair $(\textseq^{l(s)},\signseq^{s})$
and each desired source sign language $s'\ne s$, we translate the target
transcript into $l(s')$ and synthesize source signing in $s'$
(Figure~\ref{fig:bt}b):
\begin{equation}
 \label{eq:bt-construction}
 \hat{\textseq}^{l(s')}=\mathrm{MT}^{l(s)\to l(s')}(\textseq^{l(s)}),
 \qquad
 \hat{\signseq}^{s'}=\mathrm{T2S}^{s'}(\hat{\textseq}^{l(s')}).
\end{equation}
We use \textsc{\translategemma{}} 4B~\citep{translategemma2026} for MT;
$\mathrm{T2S}^{s'}$ uses a fixed generator conditioned on sign language $s'$.
The generator outputs \smplx{} motion sequences~\citep{smplx2019} rather than raw video.
Appendix~\ref{sec:appendix-implementation} gives the model and decoding
settings. Each pair
$(\hat{\signseq}^{s'},\signseq^{s})$ has a synthetic source and gold target.

Together, MT and \TS{} take the role of the reverse model in classical
back-translation. This construction requires neither a pre-existing
\StwoS{} model nor frame-level sign alignment.

Applying this procedure to the official training splits yields 112,324 ordered pairs across the six
directions among $\{\ASL,\CSL,\DGS\}$ (Table~\ref{tab:bt_stats}),
with two source languages per target corpus. Test pairs are constructed separately
(\S\ref{sec:dataset}).

Earlier sign-side back-translation constructed sign--text pairs within
one sign/spoken-language pair~\citep{zhou2021back,moryossef2021augmentation};
our construction pairs different sign languages. Keeping the target
clip gold avoids synthesizing the supervision target, but MT and \TS{}
errors can still affect the synthetic source, which may also omit
signing content absent from the transcript. We therefore evaluate
back-translated sources and existing source clips separately
(\S\ref{sec:experiment}).
\begin{table}[t]
\centering
  \caption{\textbf{\BT-constructed \StwoS training corpus (112,324 pairs).} Sources ($s'$) are synthesized from translated text; targets ($s$) are gold clips from the corresponding corpus. Each count applies to both directions sharing a target corpus. Lengths are motion timesteps (four frames, three part codes each).}
  \label{tab:bt_stats}
  \small
  \setlength{\tabcolsep}{4pt}
  \begin{tabular}{llrrr}
    \toprule
    Direction ($s' \to s$) & Gold target corpus & Pairs/direction & src\_len & tgt\_len \\
    \midrule
    \ASL $\to$ \CSL & \multirow{2}{*}{\csldaily}  & \multirow{2}{*}{18,399} & 32.37 & \multirow{2}{*}{29.27} \\
    \DGS $\to$ \CSL &                             &                         & 13.30 &                        \\
    \CSL $\to$ \ASL & \multirow{2}{*}{\howtosign} & \multirow{2}{*}{30,671} & 27.28 & \multirow{2}{*}{36.63} \\
    \DGS $\to$ \ASL &                             &                         & 15.03 &                        \\
    \ASL $\to$ \DGS & \multirow{2}{*}{\phoenix}   & \multirow{2}{*}{7,092}  & 54.43 & \multirow{2}{*}{28.29} \\
    \CSL $\to$ \DGS &                             &                         & 32.51 &                        \\
    \bottomrule
  \end{tabular}
\end{table}

\section{Joint Text-to-Sign and Sign-to-Sign Modeling}
\label{sec:model}

\begin{figure}[!htbp]
\centering
\includegraphics[width=\linewidth]{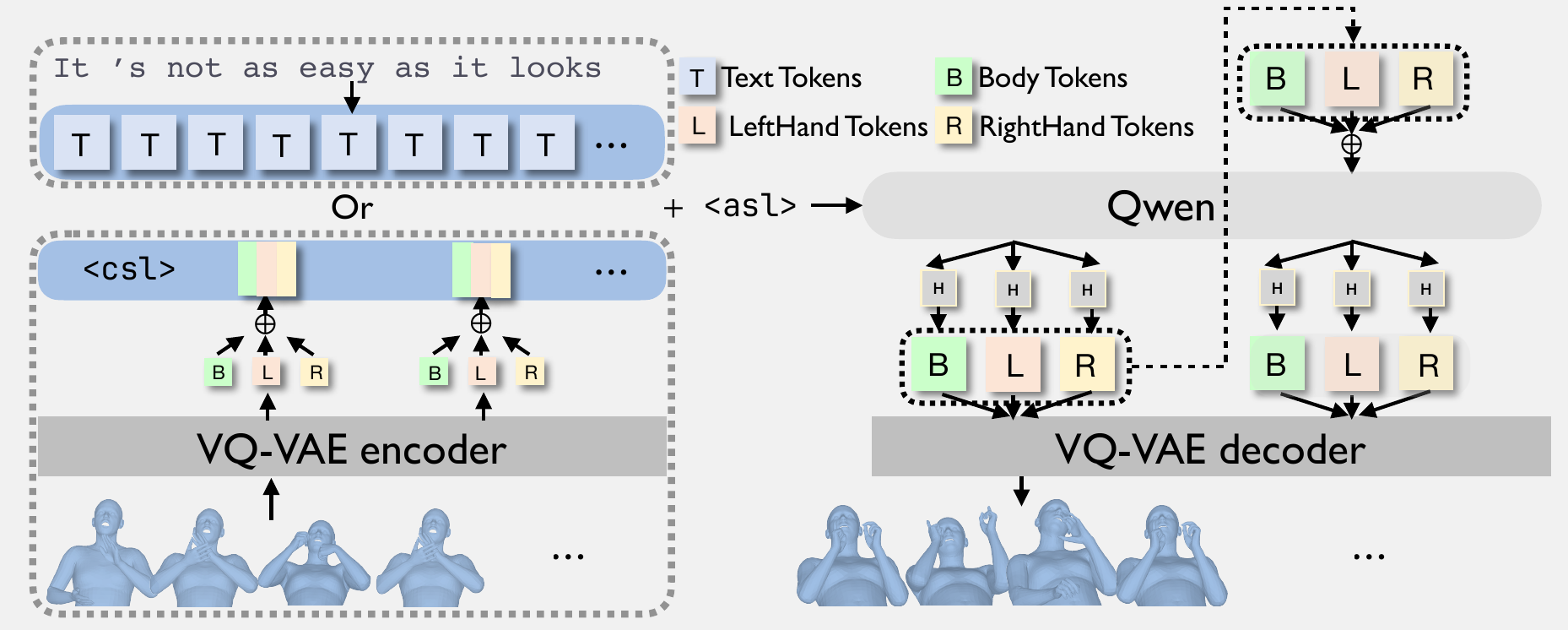}
\caption{\textbf{Shared causal model for \TS{} and \StwoS.}
Target motion is generated autoregressively without intermediate text
in the \StwoS path. Training supplies preceding gold codes.}
\label{fig:arch}
\end{figure}

We train one \qwenbackbone{} causal language model~\citep{qwen3-2025}
on natural text--sign pairs and the sign--sign pairs constructed in
\S\ref{sec:bt}. Both tasks generate target signing, conditioned on
text for \TS{} or source signing for \StwoS{}. A separately trained,
frozen VQ-VAE~\citep{vqvae2017} represents signing as discrete motion
tokens, giving the two tasks a shared output representation. The language
model predicts target tokens from the corresponding source input, and
the VQ decoder converts these predictions back to motion
(Figure~\ref{fig:arch}).

\subsection{Shared Motion Tokens and Task-Specific Prefixes}
\label{sec:model:formulation}
Each frame's 133 \smplx{} parameters~\citep{smplx2019} form a 43-channel
body stream (30 body-pose, 3 jaw, and 10 expression) and two 45-channel
hand streams.
The shared VQ-VAE encodes the three streams with separate body,
left-hand, and right-hand codebooks.
Let $\textseq^{l(s)}$ be a transcript paired with signing in language $s$.
After tokenization, we use $\signseq^s$ for the resulting motion-token
sequence and $|\signseq^s|$ for its length in
motion timesteps, including the end-of-sequence (EOS) token. A gold sign sequence is
$\signseq^{s}=(\signseq^{s}_1,\ldots,\signseq^{s}_{|\signseq^{s}|})$,
and $\hat{\signseq}^{s}$ denotes a synthetic sequence. At motion timestep
$t$, three scalar token IDs form
$\signseq^{s}_t=(y_t^{s,\mathrm{body}},y_t^{s,\mathrm{left}},y_t^{s,\mathrm{right}})$;
a non-EOS timestep represents four video frames.
We model the two translation tasks with conditional probabilities
\begin{equation}
 \label{eq:translation-tasks}
 p_{\modelparams}(\signseq^{s}\mid\textseq^{l(s)})\quad\text{and}\quad
 p_{\modelparams}(\signseq^{s}\mid\hat{\signseq}^{s'}).
\end{equation}
Here, $s'$ and $s$ denote distinct source and target sign languages,
and $\modelparams$ denotes the trainable language-model parameters.
We use natural text--sign pairs for \TS and back-translated pairs
(\S\ref{sec:bt}) for \StwoS. Both use gold targets; only the \StwoS
training source is synthetic. At inference, \StwoS can use existing
source clips.

We extend the \qwen{} vocabulary with motion-code tokens and the
\texttt{<asl>}, \texttt{<csl>}, and \texttt{<dgs>} tags. At timestep $t$,
the model's embedding table maps each token ID to a vector: the body
embedding is $\boldsymbol{e}_t^{s,\mathrm{body}}$, the left-hand embedding is
$\boldsymbol{e}_t^{s,\mathrm{left}}$, and the right-hand embedding is
$\boldsymbol{e}_t^{s,\mathrm{right}}$. We combine them into one input embedding
$\boldsymbol{e}_t^s$ using scalar hand weight $\alpha$:
\begin{equation}
 \label{eq:motion-fusion}
 \boldsymbol{e}^{s}_t=(1-2\alpha)\boldsymbol{e}_t^{s,\mathrm{body}}
             +\alpha\boldsymbol{e}_t^{s,\mathrm{left}}
             +\alpha\boldsymbol{e}_t^{s,\mathrm{right}}.
\end{equation}
We fix $\alpha=0.4$, giving weights of 0.2 for body and 0.4 for each
hand. In Figure~\ref{fig:arch}, $\oplus$ denotes this weighted fusion,
while $+$ denotes prefix concatenation.
The sequence advances once per motion timestep; text uses the
\qwen{} tokenizer and embeddings.

The \TS prefix is source text followed by the target sign-language tag.
The \StwoS prefix contains the source tag, fused source motion, an
end-of-source token, and the target tag (Figure~\ref{fig:arch}).
Each generated target position attends to the complete source prefix
and preceding target positions.

\subsection{Joint Training Objective}
\label{sec:model:multitask}
Both tasks train the same target-token predictor. With teacher forcing,
the model conditions each prediction on the source prefix and preceding
gold target tokens. Following \soke{}'s simultaneous part-wise prediction
strategy~\citep{soke2025}, we predict body, left-hand, and right-hand tokens
at each motion timestep. Our implementation uses a shared vocabulary
projection followed by three part-specific vocabulary masks and separate
softmax normalization. Each $\mathrm{H}$ in Figure~\ref{fig:arch} denotes
one such part predictor; the projection parameters are shared.
Let $\prefixseq$ be the
source prefix (text $\textseq^{l(s)}$ for \TS or synthetic signing $\hat{\signseq}^{s'}$ for \StwoS,
with language tags) and $\signseq^s$ the gold target sequence.
The conditional probability of $\signseq_t^s$ is the product of its
body, left-hand, and right-hand token probabilities, all conditioned
on the same prefix and the preceding target timesteps, denoted by
$\signseq^s_{<t}$. The sequence loss is
\begin{equation}
 \ell(\prefixseq,\signseq^s)=-\sum_{t=1}^{|\signseq^s|}
 \log p_{\modelparams}(\signseq^s_t\mid\prefixseq,\signseq^s_{<t}),
 \label{eq:token-loss}
\end{equation}
This loss sums the three parts' cross-entropy losses over time. For each task,
we sum $\ell(\prefixseq,\signseq^s)$ over the minibatch and divide by the total number of valid
target timesteps, yielding $\mathcal L_{\mathrm{T2S}}$ and
$\mathcal L_{\mathrm{S2S}}$. Source-prefix, language-tag, and padding
positions are excluded. The tasks have equal weight:
$\mathcal L=\mathcal L_{\mathrm{T2S}}+\mathcal L_{\mathrm{S2S}}.$
Section~\ref{sec:exp-ablation} compares joint and single-task training.
\subsection{Autoregressive Motion Decoding}
At inference, the target tag initializes all three streams. At each
timestep, the model selects body, left-hand, and right-hand codes, fuses
their embeddings, and appends the result as input for the next prediction.
EOS is masked in both hand streams; when the
body emits EOS, both hands are forced to emit EOS at that timestep.
The frozen VQ decoder converts output codes to motion; decoding settings
are given in Appendix~\ref{sec:appendix-implementation}.

\section{Existing Sign-to-Sign Data Set and Our High-Quality Subset}
\label{sec:dataset}

Our evaluation uses two different types of datasets:
(a) existing clips aligned across languages (see below) and 
(b) synthetic sources constructed from official held-out
text--sign pairs (\S\ref{sec:bt} and \S\ref{sec:experiment}).

\subsection{Existing Sign-to-Sign Cross-Lingual Pairs (Inan-Test)}
\label{sec:dataset:prior}
\citet{inan2025align} align clips across \ASL, \CSL, and \DGS by
translating transcripts and matching them with a multilingual paraphrase
model. Their released test set, \inantest, contains 1,781 pairs and
3,562 directed requests before availability checks
(Table~\ref{tab:cleaning_stats}). Similar transcripts need not convey
equivalent signing, so mismatched references can penalize correct
translations. We therefore complement this test set with a separate
transcript-screened diagnostic, \inancleaned.

\subsection{Extracting High-Quality Cross-Lingual Subset (Inan-Strict)}
\label{sec:dataset:cleaning}
To construct \inancleaned, we pool the prior work's released train,
development, and test pairs. For each transcript pair,
\qwenjudge{}~\citep{qwen3-2025} rates agreement
on events, actions, and key entities with an integer score from 1 to 5 and provides a brief
justification. We also compute cosine similarity between the transcripts'
multilingual MiniLM embeddings~\citep{reimers2019sbert}.
We screen for ratings of at least 4 and similarity above 0.5, then
retain selected exceptions after author discussion
(Appendix~\ref{sec:appendix-data-audit}). This screens transcript
agreement, not full linguistic equivalence of the signs.

Before availability checks, the 409 retained rows correspond to 394 distinct pairs
and 224 source-clip--target-language inputs; an input can have multiple
references. Only six pairs link ASL and DGS. Higher agreement scores
reflect selection, not independent validation of alignment.

Because Strict draws from all three partitions, it is not a subset of
\inantest. In 389 of its 409 rows, both clips occur in the underlying
corpora's training annotations. Strict therefore serves as a diagnostic
with training-clip overlap, not an unseen-clip test
(Appendix~\ref{sec:appendix-data-audit}).
\begin{table}[t]
\centering
  \caption{Inan-Test and Inan-Strict: pair counts and transcript agreement.
Strict ratings (1--5) and multilingual MiniLM-L12-v2 cosine means include
repeated pair rows. Higher ratings indicate stronger event agreement.}
  \label{tab:cleaning_stats}
  \small
  \setlength{\tabcolsep}{3pt}
  \begin{tabular}{lrrrr}
    \toprule
    & \multicolumn{2}{c}{\#Pairs} & \qwen{} rating & \sbert{} cos. \\
    \cmidrule(lr){2-3} \cmidrule(lr){4-4} \cmidrule(lr){5-5}
    Language pair & Test requested & Strict & Strict & Strict \\
    \midrule
    \ASL\bidir\CSL & 677 & 190 & 4.95 & 0.88 \\
    \ASL\bidir\DGS & 435 & 6 & 4.00 & 0.65 \\
    \CSL\bidir\DGS & 669 & 213 & 4.19 & 0.67 \\
    \midrule
    All            & 1,781 & 409 & 4.54 & 0.76 \\
    \bottomrule
  \end{tabular}
\end{table}

\begin{figure}[htbp]
  \centering
  \includegraphics[width=\linewidth]{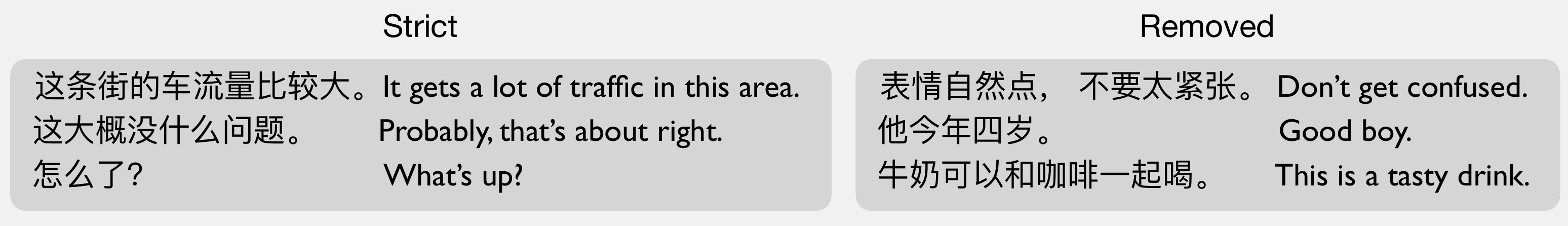}
  \caption{\textbf{Transcript pairs for \CSL--\ASL clips in the Strict
  subset.} Removed pairs are also shown for comparison.}
  \label{fig:dataset_examples}
\end{figure}

Figure~\ref{fig:dataset_examples} illustrates transcript pairs retained
in the Strict subset and removed pairs for comparison.

\vspace{-0.5cm}
\section{Experiments}
\label{sec:experiment}

We compare direct \StwoS{} with an \ST{}$\to$MT$\to$\TS{} cascade across
six directions among \ASL, \CSL, and \DGS. Inputs 
include (a) existing clips from
\inantest{} and \inancleaned{} and (b) signs back-translated from held-out
text--sign pairs (\emph{BT-input}; \S\ref{sec:bt}). The cascade's \ST{}
model depends on the input format:
\begin{itemize}
  \item \textbf{Inan-Test and Inan-Strict:}
  \unisign{}~\citep{li2025unisign} for \ASL,
  \geosign{}~\citep{fish2025geosign} for \CSL, and
  \mmslt{}~\citep{kim2025mmslt} for \DGS. The first two use pose
  inputs and the third uses RGB from the existing clips.
  \item \begin{minipage}[t]{\linewidth}
  \textbf{Back-translated source signs:}
  Generated \smplx{} motion is incompatible with these pose or RGB inputs.
  Preliminary training of alternative \ST{} models on our representation
  largely
  underperformed their published results and barely improved
  on our Sign Language Transformer (\slt{})~\citep{camgoz2020slt} baseline.
  We therefore use \slt{}, which accepts this motion, using the same model
  weights as the \slt{} evaluator for each source language.
  \end{minipage}
\end{itemize}
Both cascades use \translategemma{} 4B for MT and direct's joint
\qwenbackbone{} checkpoint and frozen sign tokenizer for \TS{}
(Appendix~\ref{sec:appendix-implementation}).

\FloatBarrier
\subsection{Evaluation Protocol and Inan-Test Comparison}
Direct and cascade share clips, references, and target-language evaluators,
though source representations can differ. \inantest{} has 3,542 common
directed requests with paired targets. Strict has 219 source-clip--target-language
anchors, with one prediction per system per anchor, scored against potentially multiple
references (availability exclusions: Appendix~\ref{sec:appendix-data-audit}).

\paragraph{Metrics}
We prioritize pose-space error: tuned pose-distance metrics show
stronger correlations with human judgments than S2T-based
BLEU~\citep{jiang2025meaningful}. Following \citet{soke2025}, we use
mean per-joint position error (\mpjpe{}) with Procrustes alignment
and FastDTW~\citep{salvador2007fastdtw}, denoted \dtwpa{}$_{\downarrow}$.
We sum joint-averaged errors in metres along the DTW path without
path-length normalization (Appendix~\ref{sec:appendix-length}). On Strict,
each system selects the reference with the lowest joint-weighted All \dtwpa{}
and uses it for all body parts. Overall means weight inputs equally.

For comparison with prior work~\citep{inan2025align}, we report
\bleufour{}$^{\uparrow}$~\citep{papineni2002bleu} on text recovered by fixed
target-language \slt{} evaluators, distinct from cascade front ends.
Corpus BLEU uses all eligible reference transcripts on Strict and paired
references on \inantest{}, without averaging directions. Gold-sign scores
(Table~\ref{tab:inan-test}) characterize evaluators on their own reference
sets; they are neither upper bounds nor a cross-paper normalization.

\begin{table}[htbp]
\centering
\caption{Direct and cascade results on \inantest{} (3,542 paired requests).
MPJPE denotes cumulative \dtwpa{} in metres; B4 denotes \bleufour{}.
Inan's scores come from \citet{inan2025align}, Tables~5--6, with different
evaluators and evaluation sets. Gold-sign evaluator \bleufour{}
(Inan / ours, each on its own reference set):
ASL 2.93 / 12.11; CSL 2.25 / 15.62; DGS 12.36 / 16.59.
Bold compares our paired systems.}
\label{tab:inan-test}
\normalsize
\setlength{\tabcolsep}{1pt}
\begin{tabular*}{\linewidth}{@{\extracolsep{\fill}}l*{12}{r}@{}}
\toprule
\multicolumn{1}{r}{\small Target} & \multicolumn{4}{c}{ASL} & \multicolumn{4}{c}{CSL} & \multicolumn{4}{c}{DGS} \\
\cmidrule(lr){2-5}\cmidrule(lr){6-9}\cmidrule(lr){10-13}
\multicolumn{1}{r}{\small Source} & \multicolumn{2}{c}{CSL} & \multicolumn{2}{c}{DGS} & \multicolumn{2}{c}{ASL} & \multicolumn{2}{c}{DGS} & \multicolumn{2}{c}{ASL} & \multicolumn{2}{c}{CSL} \\
\cmidrule(lr){2-3}\cmidrule(lr){4-5}\cmidrule(lr){6-7}\cmidrule(lr){8-9}\cmidrule(lr){10-11}\cmidrule(lr){12-13}
Method & MPJPE$_{\downarrow}$ & B4$^{\uparrow}$ & MPJPE$_{\downarrow}$ & B4$^{\uparrow}$ & MPJPE$_{\downarrow}$ & B4$^{\uparrow}$ & MPJPE$_{\downarrow}$ & B4$^{\uparrow}$ & MPJPE$_{\downarrow}$ & B4$^{\uparrow}$ & MPJPE$_{\downarrow}$ & B4$^{\uparrow}$ \\
\midrule
Inan & -- & 0.96 & -- & 0.69 & -- & 0.00 & -- & 0.00 & -- & 5.73 & -- & 5.21 \\
Cascade & \textbf{3.58} & 7.57 & 3.27 & 4.25 & 3.70 & 8.10 & 3.95 & 6.12 & 1.95 & 6.29 & 2.47 & 9.68 \\
Direct & 3.89 & \textbf{9.48} & \textbf{3.23} & \textbf{7.66} & \textbf{3.34} & \textbf{10.09} & \textbf{3.77} & \textbf{8.14} & \textbf{1.81} & \textbf{8.23} & \textbf{2.23} & \textbf{13.55} \\
\bottomrule
\end{tabular*}
\end{table}

\paragraph{Inan-Test comparison.}
With a shared evaluator, direct improves BLEU by 1.91--3.87 points
over cascade in all six directions (Table~\ref{tab:inan-test}).
Its \dtwpa{} is lower overall (3.11 versus 3.22) and in five directions.
CSL$\to$ASL is the exception, with higher errors for every part;
DGS$\to$ASL also has higher hand errors despite lower All \dtwpa{}.
Evaluator differences on gold signs prevent attributing the gap from
\citet{inan2025align} solely to translation quality.

\subsection{Direct Translation and Cascades}
\label{sec:exp-bt-input}
\label{sec:exp-cleaned}
We also compare direct and cascade on \inancleaned{} and back-translated inputs
(Tables~\ref{tab:strict-results} and~\ref{tab:main-results}).
Different examples and references prevent isolating the effect of
synthetic versus existing sources.

\begin{table}[htbp]
\centering
\caption{Direct and cascade results on \inancleaned{} (219 anchors).
Cascade \ST{} uses \unisign{}~\citep{li2025unisign} (ASL),
\geosign{}~\citep{fish2025geosign} (CSL), and
\mmslt{}~\citep{kim2025mmslt} (DGS).
The cascade uses the same MT and \TS{} models as in Table~\ref{tab:main-results}.
MPJPE denotes cumulative \dtwpa{} in metres; B4 denotes \bleufour{}.
Bold compares paired systems.
Strict overlaps training clips and has constant CSL front-end outputs; see text.}
\label{tab:strict-results}
\normalsize
\setlength{\tabcolsep}{1pt}
\begin{tabular*}{\linewidth}{@{\extracolsep{\fill}}l*{12}{r}@{}}
\toprule
\multicolumn{1}{r}{\small Target} & \multicolumn{4}{c}{ASL} & \multicolumn{4}{c}{CSL} & \multicolumn{4}{c}{DGS} \\
\cmidrule(lr){2-5}\cmidrule(lr){6-9}\cmidrule(lr){10-13}
\multicolumn{1}{r}{\small Source} & \multicolumn{2}{c}{CSL} & \multicolumn{2}{c}{DGS} & \multicolumn{2}{c}{ASL} & \multicolumn{2}{c}{DGS} & \multicolumn{2}{c}{ASL} & \multicolumn{2}{c}{CSL} \\
\cmidrule(lr){2-3}\cmidrule(lr){4-5}\cmidrule(lr){6-7}\cmidrule(lr){8-9}\cmidrule(lr){10-11}\cmidrule(lr){12-13}
Method & MPJPE$_{\downarrow}$ & B4$^{\uparrow}$ & MPJPE$_{\downarrow}$ & B4$^{\uparrow}$ & MPJPE$_{\downarrow}$ & B4$^{\uparrow}$ & MPJPE$_{\downarrow}$ & B4$^{\uparrow}$ & MPJPE$_{\downarrow}$ & B4$^{\uparrow}$ & MPJPE$_{\downarrow}$ & B4$^{\uparrow}$ \\
\midrule
Cascade & 3.71 & 8.21 & \textbf{1.59} & \textbf{5.62} & \textbf{1.40} & 8.73 & 3.21 & 7.50 & 1.57 & 4.87 & 2.36 & 8.13 \\
Direct & \textbf{2.06} & \textbf{9.87} & 2.14 & 4.52 & 2.03 & \textbf{10.96} & \textbf{2.88} & \textbf{10.25} & \textbf{1.47} & \textbf{6.79} & \textbf{1.79} & \textbf{12.66} \\
\bottomrule
\end{tabular*}
\end{table}

\begin{table}[htbp]
\centering
\caption{BT-input results (8,246 back-translated sources).
Cascade uses \slt{}~\citep{camgoz2020slt} for \ST{} and the same
MT and \TS{} models as in Table~\ref{tab:strict-results}.
MPJPE denotes cumulative \dtwpa{} in metres; B4 denotes \bleufour{}.
Bold compares paired systems.}
\label{tab:main-results}
\normalsize
\setlength{\tabcolsep}{1pt}
\begin{tabular*}{\linewidth}{@{\extracolsep{\fill}}l*{12}{r}@{}}
\toprule
\multicolumn{1}{r}{\small Target} & \multicolumn{4}{c}{ASL} & \multicolumn{4}{c}{CSL} & \multicolumn{4}{c}{DGS} \\
\cmidrule(lr){2-5}\cmidrule(lr){6-9}\cmidrule(lr){10-13}
\multicolumn{1}{r}{\small Source} & \multicolumn{2}{c}{CSL} & \multicolumn{2}{c}{DGS} & \multicolumn{2}{c}{ASL} & \multicolumn{2}{c}{DGS} & \multicolumn{2}{c}{ASL} & \multicolumn{2}{c}{CSL} \\
\cmidrule(lr){2-3}\cmidrule(lr){4-5}\cmidrule(lr){6-7}\cmidrule(lr){8-9}\cmidrule(lr){10-11}\cmidrule(lr){12-13}
Method & MPJPE$_{\downarrow}$ & B4$^{\uparrow}$ & MPJPE$_{\downarrow}$ & B4$^{\uparrow}$ & MPJPE$_{\downarrow}$ & B4$^{\uparrow}$ & MPJPE$_{\downarrow}$ & B4$^{\uparrow}$ & MPJPE$_{\downarrow}$ & B4$^{\uparrow}$ & MPJPE$_{\downarrow}$ & B4$^{\uparrow}$ \\
\midrule
Cascade & 4.35 & 7.43 & \textbf{3.92} & \textbf{8.57} & 3.39 & 8.62 & 3.50 & 8.15 & 2.63 & 9.84 & 2.57 & 10.06 \\
Direct & \textbf{3.75} & \textbf{9.65} & 4.26 & 8.43 & \textbf{3.34} & \textbf{10.81} & \textbf{3.35} & \textbf{10.19} & \textbf{2.40} & \textbf{11.62} & \textbf{2.40} & \textbf{12.28} \\
\bottomrule
\end{tabular*}
\end{table}

\paragraph{Inan-Strict results.}
On Strict (Table~\ref{tab:strict-results}), direct has lower aggregate \dtwpa{} (2.25 versus 2.44),
but is worse on ASL$\to$CSL and DGS$\to$ASL.
Direct has higher BLEU in five directions, but is lower on DGS$\to$ASL
(4.52 versus 5.62).

The 76 distinct CSL source clips have 31 reference transcripts, yet cascade
produces one front-end transcript and one motion sequence per target language.
Repeated references therefore cannot fully explain these constant outputs.
Excluding CSL sources leaves 143 anchors, where direct has higher
\dtwpa{} (2.45 versus 2.25). We retain all 219 anchors but cannot explain
the constant outputs, limiting this comparison.

\paragraph{BT-input results.}
On back-translated inputs (Table~\ref{tab:main-results}), direct improves BLEU
by 1.78--2.22 points and lowers All \dtwpa{} in five directions.
Its overall \dtwpa{} is 3.57 versus 3.70 for the cascade.
DGS$\to$ASL is the exception on both metrics: BLEU is 8.43 versus 8.57,
and \dtwpa{} is 4.26 versus 3.92.
For DGS$\to$ASL, we generate the DGS source from the English transcript via
German and compare the output with the paired gold How2Sign ASL.
Reported English--ASL misalignment~\citep{tanzer2024reconsidering}
may create source--target mismatches, though it does not alone explain the gap.

\paragraph{Direct--cascade latency.}
On 75 common inputs at batch size one, direct averages 270\,ms versus
698\,ms for the cascade's summed stage means: a $2.6\times$ speedup
(61\% lower time). Both routes are timed from prepared input representations
to generated motion, excluding loading, I/O, SMPL-X extraction, rendering,
and scoring (Appendix~\ref{sec:appendix-timing}).

\subsection{Front-End Training Data and Joint-Task Training}
\label{sec:exp-ablation}
\paragraph{S2T training data.}
We test whether the front-end training distribution affects cascade
performance on real versus synthetic inputs.
With the \slt{} architecture, MT, and \qwen{} \TS{} fixed, we compare
real, synthetic, and equal-sized mixed training data.
Synthetic training works best on synthetic inputs;
mixed training works best on real inputs
(Appendix~\ref{sec:appendix-rebuttal-evidence}). This tests \slt{}'s distribution
sensitivity, not the alternative front ends used on Inan-Test and Inan-Strict.

\paragraph{Joint versus single-task training.}
We compare the selected joint and single-task models to examine how
performance changes across \TS{} and \StwoS{} tasks.
Joint training lowers synthetic-source \StwoS{} motion error for all three
target languages relative to either single-task model, but increases \TS{}
error relative to \TS{}-only training (Table~\ref{tab:training-strategy}).
Unmatched task-specific exposure and compute prevent isolating the
training objective's causal effect
(Appendix~\ref{sec:appendix-final-joint}).

\begin{table}[htbp]
\centering
\caption{Joint versus single-task training: cumulative \dtwpa$_{\downarrow}$.
\StwoS inputs are synthetic; \TS-only performs \StwoS{} zero-shot.
Dashes mark unevaluated tasks. Training exposure is unmatched.}
\label{tab:training-strategy}
\small
\begin{tabular}{lrrrrrr}
\toprule
& \multicolumn{3}{c}{\StwoS} & \multicolumn{3}{c}{\TS} \\
\cmidrule(lr){2-4}\cmidrule(lr){5-7}
Training & \ASL & \CSL & \DGS & \ASL & \CSL & \DGS \\
\midrule
\TS only & 4.95 & 3.42 & 4.10 & 3.66 & 2.87 & 2.25 \\
\StwoS only & 4.60 & 3.62 & 2.75 & -- & -- & -- \\
Joint \TS+\StwoS & \textbf{4.00} & \textbf{3.34} & \textbf{2.40}
& 3.94 & 3.25 & 2.53 \\
\bottomrule
\end{tabular}
\end{table}

\subsection{Qualitative Analysis}
Before inspecting predictions, we select 15 examples across six directions
by sample ID and source-token length. Each shows synthetic input, direct,
cascade, and gold-reference videos with identical rendering
(Appendix~\ref{sec:appendix-qualitative}). Samples 011, 013, and 014
produce identical direct videos from different inputs; we retain these
failures. The examples do not establish semantic fidelity or
preservation of non-manual grammar and spatial reference.

\vspace{-0.4cm}
\section{Related Work}
\label{sec:related}
\vspace{-0.2cm}

\paragraph{Sign-to-text and text-to-sign translation.}
\ST{} methods include neural translation~\citep{camgoz2018slt}, joint
recognition--translation transformers~\citep{camgoz2020slt}, improved
sequence decoders~\citep{yin2020stmctransformer}, and gloss-free
models~\citep{lin2023glossfree}. \unisign{}~\citep{li2025unisign} uses pose
pretraining and optional RGB fusion; \geosign{}~\citep{fish2025geosign}
uses hyperbolic pose representations; and \mmslt{}~\citep{kim2025mmslt}
uses multimodal language-model supervision.
\TS{} methods generate signing through GANs~\citep{stoll2018slp},
progressive transformers~\citep{saunders2020progressive}, motion
primitives~\citep{saunders2021mixed}, gloss-mediated word-level
synthesis~\citep{zelinka2020neural}, or discrete tokens~\citep{soke2025}.
Combined with spoken-language MT~\citep{liu2020mbart,nllb2022,translategemma2026},
these methods form the \ST$\to$MT$\to$\TS cascade compared with our direct model.

\paragraph{Multilingual sign processing.}
MLSLT~\citep{yin2022mlslt} supports multilingual \ST{}; JWSign
~\citep{gueuwou2023jwsign} supplies multilingual data; and
WMT-SLT~\citep{muller2023wmtslt} benchmarks sign/spoken-language pairs.
\citet{jiang2023signwriting} translate into SignWriting. These efforts
connect signing with text or writing, whereas our task translates
between sign languages.

\paragraph{Cross-lingual sign-to-sign translation.}
\citet{inan2025align} align text-paired clips across \ASL, \CSL, and \DGS
to train a pose-to-pose model. We use their pairs for evaluation
(\S\ref{sec:dataset}) but construct training pairs through back-translation
(\S\ref{sec:bt}). Earlier sign-side back-translation augments sign--text
data within one sign/spoken-language pair~\citep{zhou2021back,moryossef2021augmentation};
ours pairs different sign languages. Direct translation also follows
calls for communication tools without a shared spoken-language
intermediary~\citep{yin2021including,decoster2023machine}.

\vspace{-0.2cm}
\section{Limitations}
\label{sec:limitation}
\vspace{-0.2cm}

\paragraph{Source errors and corpus coverage.}
Gold targets cannot remove MT or \TS{} errors in synthetic sources.
Although inference bypasses text, the text bridge used to construct training
pairs may lose non-manual or spatial information. Language and domain co-vary
across our corpora, limiting generalization.
We will explore OpenASL~\citep{shi2022open} and
YouTube-ASL~\citep{uthus2023youtubeasl} to assess corpus and alignment effects.

\paragraph{Data and comparison conditions.}
Strict is small, uneven, and transcript-aligned rather than naturally
parallel signing; 389 of its 409 rows overlap training clips. It does not
test unseen-clip generalization. Constant CSL front-end outputs further
limit its cascade comparison. Training controls use selected runs with
unmatched exposure rather than multi-seed estimates.

\paragraph{Metrics and evidence.}
Cumulative \dtwpa{} depends on length, \ST{}-based scoring introduces
evaluator error, and evaluator differences limit cross-paper BLEU
comparisons. Without human evaluation, automatic metrics and examples
cannot establish comprehensibility,
naturalness, or linguistic fidelity.

\paragraph{Offline inference and deployment.}
The model observes the complete source clip. Inference timing excludes
extraction and rendering and does not establish conversational latency or
deployment readiness. Partnership with the Deaf-community partnership is necessary.

\clearpage
\section*{Ethics Statement}
We plan to release code and non-restricted corpus and clip indices.
Generated data and examples will be released only where permitted by
the applicable source licenses and permissions; this release excludes
original \csldaily{} samples without authorization.
The present evaluation uses automatic metrics and qualitative examples;
it does not establish the accessibility or suitability of generated
signing for use by signing communities.
Appendix~\ref{sec:appendix-artifacts} describes the planned release
and source access conditions.

\section*{AI Use Statement}
We used OpenAI Codex to polish writing, draft manuscript sections,
find related work, and assist with research planning, evaluation code,
and result analysis. Model-assisted translation, synthetic data generation,
and transcript screening are described in Sections~\ref{sec:bt}
and~\ref{sec:dataset}. The authors reviewed the AI-assisted work and
take responsibility for the paper.
\bibliography{paper_refs}
\bibliographystyle{iclr2027_conference}
\clearpage
\appendix

\section{Supplementary Material}
\label{sec:appendix}

This appendix provides implementation settings, evaluation details,
component controls, and qualitative examples.

\subsection{Training, Decoding, and Cascade Configuration}
\label{sec:appendix-implementation}

We train a shared sign tokenizer separately and freeze it during
\TS/\StwoS training. We train the joint \qwenbackbone{} model for 100 epochs.
Training uses two GPUs, batch size 16 per GPU, and eight-step gradient
accumulation, giving a nominal effective batch size of 256. We use
seed 1234, AdamW with learning rate $2\times10^{-4}$,
decay rates $(\beta_1,\beta_2)=(0.9,0.99)$ for the first and second moments,
zero weight decay, cosine decay to $10^{-6}$,
and gradient clipping at 1.0.

Evaluation uses greedy decoding with a maximum of 100 motion-token
timesteps, test batch size 16, and seed 1234. Motion scoring uses
Procrustes alignment and FastDTW with radius 2. We report cumulative
\dtwpa{} without path-length normalization. The length analysis
in Appendix~\ref{sec:appendix-length} examines the resulting dependence
on sequence length.

\paragraph{BT source generation and text translation.}
BT sources are generated by the fixed \TS{}-only \qwenbackbone{} model
compared in \S\ref{sec:exp-ablation}, separate from the joint translation
model. It uses greedy decoding, a limit
of 100 motion-token timesteps, and earlier termination at EOS.
The intervening text translation uses
\texttt{google/translategemma-4b-it} with its official structured chat template
and language codes \texttt{zh-CN}, \texttt{en}, and \texttt{de}. Decoding is
deterministic (\texttt{do\_sample=false}, \texttt{num\_beams=1}), with
\texttt{max\_new\_tokens=128}, batch size 1, and BF16 on one GPU.

\paragraph{Language-specific sign-to-text front ends.}
For \ASL, we use the pose-only \unisign{} model with its official
\howtosign{} checkpoint. For \CSL, we use \geosign{} initialized from the
released CSL-News pretraining checkpoint and fine-tuned on \csldaily{}, with
validation-based checkpoint selection. Both use WholeBody-133 pose
inputs. For \DGS, we use RGB-based \mmslt{} on \phoenix{}. The chosen front end
is shared by the two directions with the same source language;
the downstream MT and target-sign generation stages complete the cascade.
Their standalone corpus-test \bleufour{} scores are 14.50 on \howtosign{},
23.53 on \csldaily{}, and 23.77 on \phoenix{}, respectively; these measure
source recognition, not end-to-end \StwoS{} quality.
For synthetic sources, the cascade instead uses \slt{} front ends with the
corresponding language-specific evaluation weights. Their decoding uses beam sizes
5, 5, and 3 and maximum output lengths of 100, 50, and 30 tokens for
\ASL, \CSL, and \DGS, respectively.

\subsection{Evaluation Splits, Exclusions, and Reference Scoring}
\label{sec:appendix-data-audit}

The 112,324 training pairs use the official training splits of
\howtosign{}, \csldaily{}, and \phoenix{}; BT-input evaluation uses their
official test splits. Pairing is at the utterance level, with synthetic
sources and gold targets as described in \S\ref{sec:bt}.

\inantest{} requests 3,562 directed translations. Direct produces 3,542
outputs, with 20 requests lacking source or target motion assets. The
cascade produces all 3,562 outputs, of which 3,552 have target motion for
scoring. The paired Inan-Test tables compare both systems on the same
3,542 requests, scoring each against its paired target reference.

Strict independently pools the prior work's train, development, and test partitions.
Of its 409 rows, 389 have both clips in monolingual training annotations,
one in development, and 19 in test. It is therefore a diagnostic on existing clips, not an unseen-clip test.

Four rows with \qwen{} ratings below 4 and 18 with cosine similarity at or
below 0.5 were retained after author discussion. The fixed manifest
contains 394 distinct clip pairs; 15 occur twice. Repeated pair rows do
not add generation trials or references. Before availability checks, the six directions contain 24, 6, 70, 66, 6,
and 52 anchors, respectively, in the column order of Table~\ref{tab:strict-results}.

Four three-frame ASL clips cannot be loaded by the fixed evaluation
pipeline and are excluded as inputs and references, affecting ten
manifest rows. We also exclude one ASL$\to$DGS source absent from
\unisign{}'s released inputs from both systems' comparison, retaining
the same clip as a DGS$\to$ASL reference. This leaves 219 anchors and
767 source--reference comparisons; the 409-row inventory is preserved.

\paragraph{Multi-reference selection and sample-weighted aggregation.}
Each source anchor produces one prediction. Both systems use the same
candidate references, deduplicated by target clip. For each candidate,
we compute cumulative \dtwpa using a separate alignment path for each
part. Let $C_{\mathrm{body}}$ denote the body error,
$C_{\mathrm{left}}$ the left-hand error, and $C_{\mathrm{right}}$ the
right-hand error. Weighting by 12 body joints and 21 joints per hand gives
the aggregate $C_{\mathrm{all}}$:
\begin{equation}
  C_{\mathrm{all}}=
  \frac{12C_{\mathrm{body}}+21C_{\mathrm{left}}+21C_{\mathrm{right}}}{54}.
  \label{eq:strict-part-cost}
\end{equation}
Each system selects the reference minimizing $C_{\mathrm{all}}$ and uses
that same reference for all reported parts; ties are broken by target
clip ID. No path-length normalization is applied. Overall motion Avg
weights directions by their evaluated anchor counts.

\bleufour{} uses all eligible reference transcripts in a corpus calculation
for each direction. Hypotheses are recovered from generated motion by
the same target-language evaluator for both systems; the cascade's
intermediate text is not used for scoring. No overall BLEU is reported.
The ASL$\to$DGS and DGS$\to$ASL directions have only five and six anchors,
respectively, and anchors
sharing references need not be statistically independent.

Table~\ref{tab:strict-direct-parts} reports both systems on this common subset.

\begin{table}[!htbp]
  \centering
  \caption{Direct and cascade motion errors by body part on Inan-Strict. The CSL-source cascade front end produces constant text (see main text).
  All entries use cumulative \dtwpa$_{\downarrow}$; the body and both
  hands use the same selected reference for each prediction. Avg
  gives each evaluated anchor equal weight.}
  \label{tab:strict-direct-parts}
  \small
  \begin{tabular}{llrrrrr}
    \toprule
    Method & Direction & Anchors & Body & Left hand & Right hand & All \\
    \midrule
    \multirow{7}{*}{Direct}
    & ASL$\to$CSL & 66 & 5.60 & 0.48 & 1.55 & 2.03 \\
    & ASL$\to$DGS & 5 & 3.97 & 0.69 & 0.81 & 1.47 \\
    & CSL$\to$ASL & 24 & 4.53 & 1.35 & 1.37 & 2.06 \\
    & CSL$\to$DGS & 52 & 4.69 & 0.76 & 1.16 & 1.79 \\
    & DGS$\to$ASL & 6 & 6.78 & 2.12 & 2.06 & 2.14 \\
    & DGS$\to$CSL & 66 & 7.68 & 1.11 & 1.91 & 2.88 \\
    \cmidrule(lr){2-7}
    & Avg & 219 & 5.89 & 0.88 & 1.54 & 2.25 \\
    \midrule
    \multirow{7}{*}{Cascade}
    & ASL$\to$CSL & 66 & 3.86 & 0.30 & 1.10 & 1.40 \\
    & ASL$\to$DGS & 5 & 3.53 & 0.95 & 1.08 & 1.57 \\
    & CSL$\to$ASL & 24 & 7.84 & 2.39 & 2.68 & 3.71 \\
    & CSL$\to$DGS & 52 & 5.46 & 1.55 & 1.41 & 2.36 \\
    & DGS$\to$ASL & 6 & 3.42 & 1.05 & 1.08 & 1.59 \\
    & DGS$\to$CSL & 66 & 8.81 & 1.14 & 2.09 & 3.21 \\
    \cmidrule(lr){2-7}
    & Avg & 219 & 6.15 & 1.11 & 1.64 & 2.44 \\
    \bottomrule
  \end{tabular}
\end{table}

\subsection{Front-End Training, Joint-Model Results, and Tokenizer Errors}
\label{sec:appendix-rebuttal-evidence}

The following tables report auxiliary control experiments. System identities
are given explicitly because these controls address different questions from
the direct--cascade comparison on Inan-Test and Inan-Strict.

\subsubsection{Real, Synthetic, and Mixed Front-End Training}

To test sensitivity to the source-sign distribution, we keep the \slt{}
front-end architecture, MT model, and \qwen{}-based \TS{} model fixed.
We train the front end on real signs, \TS{}-generated signs, or an equal-sized
50/50 mixture (Table~\ref{tab:appendix-distribution-control}). This control
does not compare the \unisign{}, \geosign{}, or \mmslt{} front ends used
for Inan-Test and Inan-Strict.

\begin{table}[!htbp]
  \centering
  \caption{Real, synthetic, and mixed training data for the \slt{} front end, with \qwen{}-based
  \TS{}. These auxiliary reported aggregates cover six directions and are separate from the paired cohorts in the main tables. \dtwpa is lower-is-better; B4 is higher-is-better.}
  \label{tab:appendix-distribution-control}
  \small
  \setlength{\tabcolsep}{7pt}
  \begin{tabular}{lrrrr}
    \toprule
    & \multicolumn{2}{c}{Real test inputs} &
      \multicolumn{2}{c}{Synthetic test inputs} \\
    \cmidrule(lr){2-3}\cmidrule(lr){4-5}
    \ST{} training inputs & \dtwpa & B4 & \dtwpa & B4 \\
    \midrule
    Real & 8.23 & 7.26 & 8.20 & 7.18 \\
    Synthetic & 8.49 & 6.86 & 7.78 & 7.85 \\
    Equal-sized 50/50 mixture & 8.14 & 7.37 & 7.92 & 7.61 \\
    \bottomrule
  \end{tabular}
\end{table}

On synthetic test inputs, training the front end on synthetic signs gives
lower motion error and higher BLEU than training on real signs. On real
test inputs, the mixed-training model has the lowest motion error and
highest BLEU among these three settings. The preferred training distribution
therefore depends on the test inputs; these controls do not establish that
source-distribution mismatch is absent.

\subsubsection{Joint-Model Motion and BLEU by Task and Target Language}
\label{sec:appendix-final-joint}

Table~\ref{tab:appendix-final-joint} reports the selected \qwenbackbone{}
checkpoint trained jointly on \TS and \StwoS. Evaluation covers 4,126
held-out \TS{} text--sign inputs and 8,246 matched synthetic-source
\StwoS{} inputs with gold target signing. Target-language and overall
motion aggregates weight their contributing inputs equally. Each \StwoS{}
target-language row pools the two source languages; the six directional
results appear in Table~\ref{tab:main-results}. ``All'' is the evaluator's
reported aggregate, rather than an unweighted mean of the three displayed
parts. \StwoS{} BLEU is reported by direction, without averaging two corpus
scores into a target-language score.

For the joint-versus-single-task comparison in
Table~\ref{tab:training-strategy}, the \TS{}-only and joint runs use a
cosine learning-rate floor of $10^{-6}$, while \StwoS{}-only uses $10^{-7}$.
All three runs use a nominal effective batch size of 256, but their selected
checkpoints are not matched for task-specific exposure or compute.

\begin{table}[!htbp]
  \centering
  \caption{Joint \TS/\StwoS model: cumulative \dtwpa$_{\downarrow}$ and \bleufour{}$^{\uparrow}$, separated by task and target
  language. $N$ counts the evaluation inputs. Target-pooled \StwoS
  BLEU is not reported; its per-direction scores appear in
  Table~\ref{tab:main-results}.}
  \label{tab:appendix-final-joint}
  \small
  \setlength{\tabcolsep}{3pt}
  \begin{tabular}{lllrrrrrr}
    \toprule
    Method & Task & Target & Body & Left hand & Right hand & All & $N$ & \bleufour{} \\
    \midrule
    \multirow{6}{*}{Joint}
    & \TS & \ASL & 9.02 & 2.42 & 2.56 & 3.94 & 2308 & 8.08 \\
    & \TS & \CSL & 8.24 & 1.54 & 2.10 & 3.25 & 1176 & 9.04 \\
    & \TS & \DGS & 6.35 & 1.34 & 1.54 & 2.53 & 642 & 10.52 \\
    \cmidrule(lr){2-9}
    & \StwoS & \ASL & 9.03 & 2.49 & 2.64 & 4.00 & 4610 & -- \\
    & \StwoS & \CSL & 8.44 & 1.50 & 2.28 & 3.34 & 2352 & -- \\
    & \StwoS & \DGS & 6.09 & 1.16 & 1.54 & 2.40 & 1284 & -- \\
    \bottomrule
  \end{tabular}
\end{table}

Compared with the cascade on synthetic-source inputs (Table~\ref{tab:main-results}),
Direct has higher errors for all parts in DGS$\to$ASL;
its right-hand error is also higher for ASL$\to$CSL (2.29 versus 2.20)
and marginally for CSL$\to$DGS (a difference below 0.01).

\subsubsection{Shared VQ-VAE Reconstruction Error}

Reconstruction measures the tokenizer's ability to represent gold
motion; it does not directly establish the semantic fidelity of generated
signing.

Our tokenizer uses one shared VQ-VAE with separate body, left-hand, and
right-hand codebooks, rather than three VQ-VAEs as in \soke~\citep{soke2025}.
The parts have 43/45/45 input channels and 96/192/192 codebook entries,
with 512-dimensional embeddings and four-fold temporal downsampling.
We use CSL-derived mean and standard deviation for normalization.
The pooled token corpus uses all 96/192/192 codes; this distinct-code
coverage does not imply uniform usage or full coverage within every language.

Tokenizer training uses one GPU, batch size 320, a 500-epoch schedule,
and AdamW with learning rate $2\times10^{-4}$, moment decay rates
$(\beta_1,\beta_2)=(0.9,0.99)$,
zero weight decay, and cosine decay to $10^{-6}$.
Table~\ref{tab:appendix-vq-reconstruction} reports the shared tokenizer's
frame-aligned \pampjpe{}, without DTW.

\begin{table}[!htbp]
  \centering
  \caption{Shared VQ-VAE reconstruction errors by language and body part.
  Values are frame-aligned \pampjpe{}$_{\downarrow}$ in millimetres, without DTW.
  ``Hand'' averages the two hands; ``All'' weights the 14 body joints
  and 21 joints per hand equally.}
  \label{tab:appendix-vq-reconstruction}
  \small
  \setlength{\tabcolsep}{3pt}
  \begin{tabular}{lrrrrrrrrr}
    \toprule
    & \multicolumn{3}{c}{\ASL} & \multicolumn{3}{c}{\CSL} & \multicolumn{3}{c}{\DGS} \\
    \cmidrule(lr){2-4}\cmidrule(lr){5-7}\cmidrule(lr){8-10}
    Method & All & Body & Hand & All & Body & Hand & All & Body & Hand \\
    \midrule
    Shared VQ-VAE & 9.83 & 19.17 & 6.72 & 9.69 & 22.84 & 5.31 & 11.31 & 24.75 & 6.83 \\
    \bottomrule
  \end{tabular}
\end{table}

\subsection{Reference-Length Dependence of Cumulative \dtwpa{}}
\label{sec:appendix-length}

For the \qwenbackbone{} \TS-only checkpoint, the saved evaluator inputs
provide reference lengths after evaluator preprocessing: 40--400 frames,
rounded to multiples of four. Cumulative \dtwpa{} depends on both local
joint errors and alignment-path length. Dividing \dtwpa{} by reference
length is a sensitivity
analysis, not exact path normalization.

\begin{table}[!htbp]
  \centering
  \caption{Spearman rank correlation $\rho$ between evaluator reference-frame length and
  cumulative \dtwpa, before and after division by reference
  length. Brackets show percentile 95\% intervals from 3,000
  bootstrap resamples of evaluation examples.}
  \label{tab:appendix-qwen-length}
  \small
  \begin{tabular}{lrrr}
    \toprule
    Language/dataset & $N$ & \dtwpa{} $\rho$ &
      Reference-normalized $\rho$ \\
    \midrule
    ASL/\howtosign{} & 2,308 & 0.935 [0.924, 0.945] &
      $-0.160$ [$-0.200$, $-0.120$] \\
    CSL/\csldaily{} & 1,176 & 0.863 [0.844, 0.879] &
      $-0.201$ [$-0.256$, $-0.141$] \\
    DGS/\phoenix{} & 642 & 0.928 [0.914, 0.940] &
      $-0.081$ [$-0.164$, $-0.003$] \\
    \bottomrule
  \end{tabular}
\end{table}

After reference-length normalization, all three correlations are small and
negative. Thus, these saved outputs provide no evidence that the score divided
by reference length increases with longer reference sequences. They do not show
that \qwen{} \StwoS{} quality is length-invariant.  Generated lengths and DTW path
lengths were not saved, so exact path-normalized error is unavailable
from these saved scores.

\subsection{Fixed-Case Selection and Video Rendering}
\label{sec:appendix-qualitative}

We select examples before inspecting predictions: the lexicographically
first sample and the shortest and longest source-token sequences in
each direction, with sample-ID tie breaking and deduplication. This
produces 15 examples, each with synthetic input, direct prediction,
cascade prediction, and gold reference videos (60 videos in total).
The supplementary package includes 56 videos, withholding the four
gold references from original \csldaily{} samples. Generated videos
are retained for all 15 cases.

Videos share a camera and renderer, retain full sequences, and use each
dataset's nominal frame rate without smoothing or temporal alignment.
Contact sheets independently show first/middle/last frames; reference
panels use the evaluator's 40--400-frame preprocessing, without
synchronization across panels. \slt{} readouts are model predictions,
not human annotations.

\subsection{Stage Timings and Batch-Size Effects}
\label{sec:appendix-timing}

Measurements of direct \StwoS{} and the cascade's \TS{} stage use an H200 NVL MIG 3g.71gb
partition, FP32 without autocast, batch size one, greedy decoding capped
at 100 timesteps, no KV cache, five warm-up batches, and three repeats.
The cohort selects the first 16 canonical anchors per direction, or all
available anchors when fewer exist, then applies the source-availability
exclusion described in Appendix~\ref{sec:appendix-data-audit}.
This leaves 75 anchors and 225 timing measurements, weighted equally.

Direct averages 270.293\,ms: 2.772\,ms source VQ encoding,
264.654\,ms autoregressive generation, and 2.867\,ms target decoding and
motion assembly. It starts with device-resident pose and excludes CPU
preprocessing and host-to-device transfer. Cascade \TS{} averages
380.508\,ms from in-memory translated text, including prompt construction,
tokenization, host-to-device transfer, generation, and target decoding.
It excludes source recognition and MT. Both exclude model loading,
external I/O, SMPL-X extraction, rendering, and scoring.

The \ST{} and MT measurements use the same 75 inputs and three repeats
at batch size one, averaging 117.499 and 200.090\,ms. Adding the \TS{}
mean gives 698.097\,ms, a $2.58\times$ speedup for direct and a 61.3\%
reduction in mean inference time. This total sums separately timed stages;
it is not a single uninterrupted pipeline measurement. The comparison
starts from each route's prepared input representation, including RGB
for \mmslt{}, and ends at generated motion, excluding extraction and rendering.

Batch-size-one cascade outputs differ from the formal batched outputs on
two of 75 anchors: one produces 156 rather than 188 frames; the other
has different motion values at the same 28-frame length. The previously
excluded direct anchor produced 32 versus 144 frames, but its exclusion
was based on source-input availability. Timing therefore does not establish
identical outputs across batch sizes.

\subsection{Planned Artifacts and Dataset Release Restrictions}
\label{sec:appendix-artifacts}

We plan to release code, model configurations, and non-restricted
corpus/clip indices. Generated cross-lingual data, evaluation inputs
and references, and qualitative examples will be released only where
permitted by the applicable source terms. We have not obtained
authorization to redistribute original \csldaily{} samples and exclude
them from the planned public release.

\paragraph{Dataset access and redistribution terms.}
\howtosign{} is distributed under CC BY-NC 4.0 with attribution and a
non-commercial restriction.\footnote{\href{https://how2sign.github.io/}{How2Sign dataset license}}
\csldaily{} access requires a research agreement; its terms prohibit
redistributing the original dataset without prior written
consent.\footnote{\href{https://ustc-slr.github.io/datasets/2021_csl_daily/Release-Agreement-CSL-Daily.pdf}{CSL-Daily release agreement}}
Original video clips will be referenced through indices and their source
access channels. Releases derived from \howtosign{} and \phoenix{} will
follow their respective licenses and access terms, including applicable
attribution and non-commercial restrictions. Public download access
alone is not treated as permission to redistribute source material.

\end{document}